\documentclass[runningheads]{llncs}
\usepackage[T1]{fontenc}
\usepackage{graphicx}
\usepackage[utf8]{inputenc} % allow utf-8 input
\usepackage[T1]{fontenc}    % use 8-bit T1 fonts
\usepackage{hyperref}       % hyperlinks
\usepackage{url}            % simple URL typesetting
\usepackage{booktabs}       % professional-quality tables
\usepackage{amsfonts}       % blackboard math symbols
\usepackage{nicefrac}       % compact symbols for 1/2, etc.
\usepackage{microtype}      % microtypography
\usepackage{xcolor}         % colors
\usepackage{amsmath} 
\usepackage{multirow}
\usepackage{booktabs,arydshln}
\usepackage{subcaption}
\usepackage[colorinlistoftodos,bordercolor=orange,backgroundcolor=orange!20,linecolor=orange,textsize=scriptsize]{todonotes}
\makeatletter
\def\adl@drawiv#1#2#3{%
        \hskip.5\tabcolsep
        \xleaders#3{#2.5\@tempdimb #1{1}#2.5\@tempdimb}%
                #2\z@ plus1fil minus1fil\relax}
\newcommand{\cdashlinel}[1]{%
  \noalign{\vskip\aboverulesep
           \global\let\@dashdrawstore\adl@draw
           \global\let\adl@draw\adl@drawiv}
  \cdashline{#1}
  \noalign{\global\let\adl@draw\@dashdrawstore
           \vskip\belowrulesep}}
\makeatother

\begin{document}
\title{BitNet b1.58 Reloaded: State-of-the-art Performance Also on Smaller Networks}
\titlerunning{BitNet b1.58 Reloaded}
\author{Jacob Nielsen\inst{1}\orcidID{0009-0009-8141-630X} \and\\
Peter Schneider-Kamp\inst{1}\orcidID{0000-0003-4000-5570}}
\authorrunning{Nielsen \& Schneider-Kamp}
\institute{Department of Mathematics and Computer Science, University of Southern Denmark, Odense, Denmark\\
\email{\{jacn,petersk\}@imada.sdu.dk}}
\maketitle              % typeset the header of the contribution
\begin{abstract}
% The abstract should briefly summarize the contents of the paper in
Recently proposed methods for 1-bit and 1.58-bit quantization aware training investigate the performance and behavior of these methods in the context of large language models, finding state-of-the-art performance for models with more than 3B parameters. In this work, we investigate 1.58-bit quantization for small language and vision models ranging from 100K to 48M parameters. We introduce a variant of BitNet b1.58, which allows to rely on the median rather than the mean in the quantization process.
Through extensive experiments we investigate the performance of 1.58-bit models obtained through quantization aware training. We further investigate the robustness of 1.58-bit quantization-aware training to changes in the learning rate and regularization through weight decay, finding different patterns for small language and vision models than previously reported for large language models.
Our results showcase that 1.58-bit quantization-aware training provides state-of-the-art performance for small language models when doubling hidden layer sizes and reaches or even surpasses state-of-the-art performance for small vision models of identical size. Ultimately, we demonstrate that 1.58-bit quantization-aware training is a viable and promising approach also for training smaller deep learning networks, facilitating deployment of such models in low-resource use-cases and encouraging future research.

\end{abstract}

\section{Introduction}
The recent years of development of natural language processing (NLP) have been dominated by the capabilities offered by Large Language Models (LLMs). However, due to the size of these models, they pose a challenge in deployment and raise concerns regarding the environmental impact. Post-training quantisation methods transform the 16-bit weights to a lower bit-representation, which both reduces the memory and computational needs. The idea is to take the trained weights and find a good way of mapping them to fewer bits, enabling more efficient inference.

Several post-training quantisation methods have been proposed, including but not limited to, Generative Pre-trained Transformer Quantization \cite{frantar2023gptq} and Activation-aware Weight Quantization \cite{lin2024awq}. However, post-training quantization inherently comes at the cost of precision. Post-training quantization has also been employed in other domain such as vision models \cite{li2023vit}.

An alternative to post-training quantization is quantization-aware training such as LLM-QAT \cite{liu2023llmqat} and QA-LoRA. Here, as the training optimizes the quantized weights, there is no loss of precision when using the quantized model for inference.  Recent works on 1-bit \cite{wang2023bitnet} and 1.58-bit \cite{ma2024era} quantization-aware training architectures have demonstrated the potential of training in very low-bit representation while still maintaining most or all of the performance for LLMs.

The 1.58-bit quantization aware training architecture BitNet b1.58\cite{ma2024era} proposes a solution based on replacing linear 16-bit layers with layers where the weights only assume the values $-1$, $0$, and $1$. Notably, for large-enough LLMs, BitNet b1.58 can match the 16-bit precision baselines both in capacity and performance. From above 3B parameters, the 1.58-bit models trained from scratch perform just as well as 16-bit models.

In this work we investigate 1.58-bit quantization aware training for small language models (SLMs) and vision models ranging from 100K to 48M parameters. We introduce a variant of BitNet b1.58 that relies on the median rather than the mean of the absolute values of the weights. Through extensive experiments we investigate and compare the scaling, the learning-rate robustness, and the regularization properties of both 1.58-bit variants. Our work demonstrates that 1.58-bit quantization aware training can get close to state-of-the-art performance on SLMs and even exceed the state-of-the-art performance on vision models, opening a new avenue for research in this direction. This facilitates the deployment of SLMs and small vision models in low-ressource settings. Our implementation is available from GitHub\footnote{https://github.com/schneiderkamplab/bitlinear} and the Python Packacking Index\footnote{https://pypi.org/project/bitlinear/}.

% Using the proposed method on small models pushes the gradients to high magnitudes effectively preventing the network to learn. We refer to this as "weight implosion". In this work, we propose a modifications to \cite{158bit}'s method introducing the capabilities of the powefull 1-Bit scheme to small language models (SLM) and vision models for classification. \\

% document the problem:
% \begin{itemize}
%     \item On tiny mistral models: first a bit of extra loss, then the gradients explode -> coursing weight implosion
% \end{itemize}

% \subsubsection{Why does 1.58 architecture work?}
% \jacob{I am not sure if we should have such a section, just part of it or just for me to understand the mechanisms}
% \begin{enumerate}
%     \item high precision shadow weights 
%     \item high precision activation bits 
%     \item STE
%     \item moved capacity from weight resolution to pattern of the ternary system
%         \item capacity moved from 16bit resolution, to utilize the models parametres. Prior work demonstrates that ~30\% of the weights does not participate in performance gain.
%     \item quantise weight for inference to speed up.
% \end{enumerate}

\begin{figure}[t]
\centering
\includegraphics[width=1.0\textwidth]{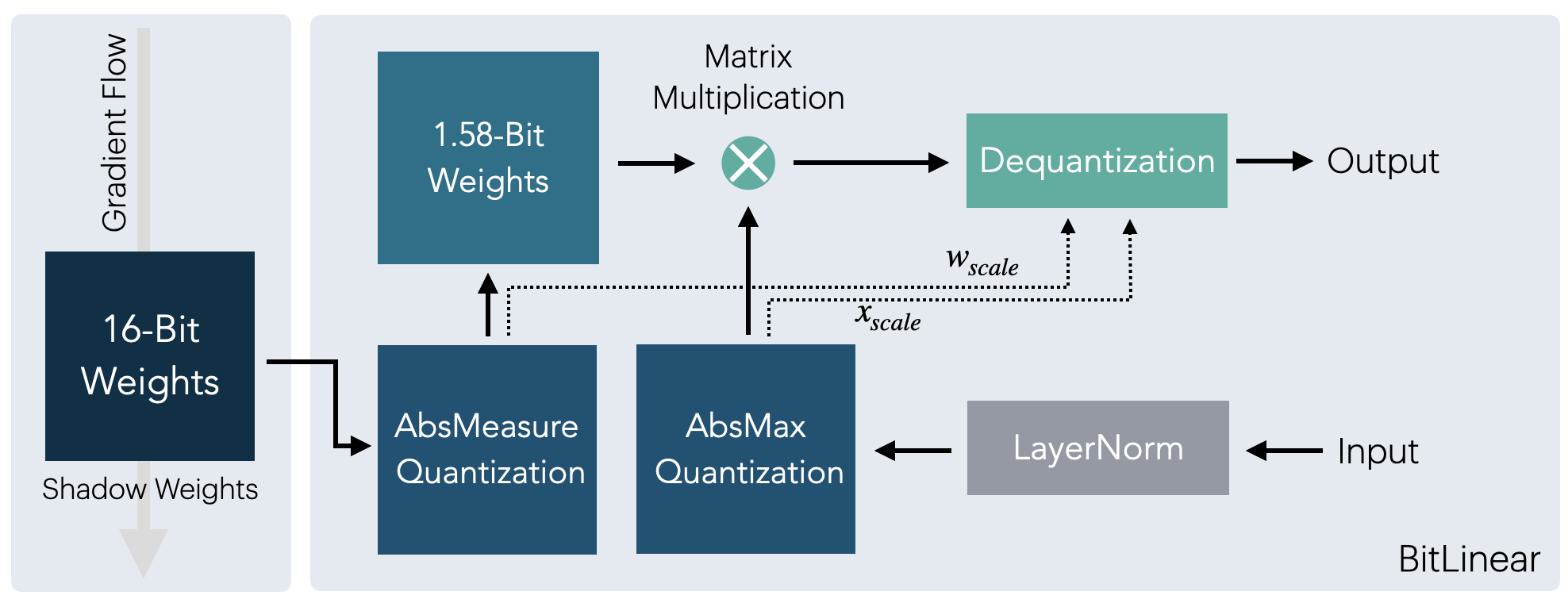}
% \vspace{-3pt}
\caption{The \texttt{BitLinear} layer is the backbone of the BitNet 1.58 Bits Reloaded architecture. It provides a drop-in replacement for linear layers (often referred to as feed-forward networks or multi-level perceptrons) in any architecture. \emph{AbsMeasure} denotes the \texttt{mean} or \texttt{median} of the absolute values of the weight. The two factors $x_{scale}$ and $w_{scale}$ denote two scaling factors for the input and 16-bit weights respectively, used in the dequantization. We employ a straight-through estimator for the backward computations of the gradients.}
\label{fig:bitlinear_flow_2}
\vspace{-2ex}
\end{figure}

\section{Method}\label{sec:method}
In this section we present our quantization aware training architecture as a generalization of the BitNet b1.58 architecture \cite{ma2024era}. First, we present our quantization method. Then, we document our experimental setup.

\subsection{b1.58 Quantization}
Our \texttt{BitLinear} layer functions as a drop-in replacement for PyTorch's \texttt{torch.nn.Linear} layer. Figure \ref{fig:bitlinear_flow_2} illustrates \texttt{BitLinear}'s 5-step computation flow:
\begin{enumerate}
    \item The activations are normalized.
    \item The normalized activations are quantized to k-bit precision.
    \item The 16-bit shadow weights are quantized to 1.58-bit weights.
    \item The quantized activations are multiplied with the 1.58-bit weights.
    \item The result of the multiplication is dequantized by rescaling.
\end{enumerate}

\noindent In the following, we details the mathematics behind this computation flow. We denote the Layer normalization~\cite{ba2016layer} of input $I$, as $\hat{I}$. We then define the quantified activation-bits as $x_{scale}$, constituting the \textsf{AbsMax}: % should be x_scale?
\begin{equation}\label{input_scale_activation_bits}
    x_{scale} = \frac{Q_b}{max(|\hat{I}|) + \epsilon} 
\end{equation}
where $Q_b = 2^{k-1}$ is the range of the k bits used for the quantized activation.  $\epsilon$ is s small value preventing zero-division. This means all activations can be scaled to integer values $\{-Q_b-1, \ldots, Q_b\}$. We define the \textsf{AbsMax Quantization} for the activations as follows:
\begin{equation}
    x_{quant} = max(-Q_B, min(Q_B-1, round(\hat{I} \cdot x_{scale}))
\end{equation}
 Furthermore, we quantize the 16-bit weights $W \in \mathcal{R}^{n\times m}$ to a ternary system of integer values $\{-1, 0, 1\}$ as follows. We define the scaling of $W$ as:
\begin{equation}
    w_{scale} = \frac{1}{Measure(|W|) + \epsilon}
\end{equation}
Where $Measure$ denotes either the mean or median function, constituting the \textsf{AbsMeasure Quantization}.

We define the quantized weights $W_{quant}$ (denoted as \textsf{1.58-Bit Weights} in Figure~\ref{fig:bitlinear_flow_2}) as: % same as for activations
\begin{equation}
   W_{quant} = max(-1, min(1, round(W \cdot w_{scale}))
\end{equation}
Having quantized both the activations and the weights, we can apply a kernel with $q_{quant}$ and $w_{quant}$ as inputs: 
\begin{equation}
   y_{quant} = x_{quant} \cdot W_{quant} + b
\end{equation}
where $b$ is optional bias. We detach both $x_{quant}$ and $w_{quant}$ from the computation graph to achieve a straight-through estimation of the gradients. The gradients update the ``shadow weights'', i.e., the \textsf{16-Bit Weights} that are quantized by \textsf{AbsMeasure Quantization}.

Finally, we rescale the output $y$ during the \textsf{Dequantization} process:
\begin{equation}
  y= \frac{y_{quant}}{w_{scale} \cdot x_{scale}}
\end{equation}

Comparing to the original BitNet b1.58, there are a number of differences:
\begin{itemize}
    \item We chose to use a standard layer normalization (\textsf{LayerNorm}) rather than RMS normalization, as the computational overhead is minimal and we observed slightly better performance with the standard layer norm in preliminary experiments.
\item We allow the use of both the median and the mean for quantizing weights. Prior works \cite{wang2023bitnet,ma2024era} solely employ the mean. We investigate the impact of this choice in Section~\ref{sec:results}.
    \item We actually quantize weights and activations to integer values. This means the matrix multiplications are performed between the 1.58-bit weights with integer values $\{-1, 0, 1\}$ and the 8-bit quantized activations with integer values ${-128, \ldots, 127}$. This allows to develop multiplication-free kernels, as multiplication with $-1$ corresponds to the subtraction of an 8-bit integer value, multiplication with $0$ to the disregard of a value, and multiplication with $1$ to the addition of an 8-bit integer value.

    This is in contract to previous work \cite{ma2024era}, where the quantized weights have floating point values $\{\frac{-1}{w_{scale}}, 0, \frac{1}{w_{scale}}\}$ while quantized activations have floating point values $\{\frac{-128}{x_{scale}}, \ldots, \frac{127}{x_{scale}}\}$ according to the published information about the implementation\cite{158bit_tips_code}. Consequently, our BitNet 1.58 Bits Reloaded architecture is more directly amenable to custom software kernels and hardware implementations.
\end{itemize}

% \subsection{Implementation details}
% 1)
% Experiments for small vision models are conducted with the \texttt{Adam}\cite{kingma2014adam} optimizer and with a batch-size of 128. The number of model parameters is slightly higher in the BitLinear setting, as we both have 1.58-bit weights as well as the 16-bit shadow weights. However, this fact does not change the number of trainable/optimized parameters in practice.  

% 2)The experiments for SLMs are conducted with the standard trainer from the Hugging Face transformers library\footnote{https://github.com/huggingface/transformers}.

% 3) The experiments for vision models in Table \ref{tab:vision_baselines} are based on Pytorch Lightning\footnote{https://github.com/Lightning-AI/pytorch-lightning} and use torchvision's\footnote{https://pytorch.org/vision/stable/index.html} versions of the datasets.

\subsection{Experimental setup}
We conduct all experiments with standard networks in small configurations with the \texttt{torch.nn.Linear} layers replaced by our \texttt{BitLinear} layers. The \texttt{Adam}\cite{kingma2014adam} optimizer and a batch-size of 128 are employed. The number of model parameters is slightly higher in the BitLinear setting, as we both have 1.58-bit weights as well as the 16-bit shadow weights. However, this fact does not change the number of trainable/optimized parameters in practice. 

For SLMs, we train small Mistral-like models with 4 layers and hidden sizes of $32$, $64$, $128$, and $256$. The number of attention head and key-value cache heads is set to the ceiling of the hidden size divided by 64, i.e., $1$ head for $32$ and $64$ hidden sizes and $2$ and $4$ heads for $128$ and $256$, respectively. The resulting models sizes are 6M, 12M, 24M, and 48M parameters. We use a text corpus of 135M tokens and train from scratch for 10 epochs unless otherwise noted, corresponding to a total of 1.35B tokens for each training. We trained a Byte Pair Encoding tokenizer with a vocabulary size of $8{,}000$. The experiments are conducted with the standard trainer from the Hugging Face transformers library\footnote{https://github.com/huggingface/transformers}.

For vision models, we consider a standard serial implementation of classifier for \texttt{MNIST} and standard CNN-based implementations for \texttt{CIFAR-10}, and \texttt{CIFAR-100}. The model for \texttt{MNIST} is the smallest in this paper with only 100K parameters. The \texttt{CIFAR-10} and \texttt{CIFAR-100} models represent the but smallest models with 2.1M and 2.2M, respectively. The difference in model size is explained by \texttt{CIFAR-10} having 10 classes and \texttt{CIFAR-100} having 100 classes. The experiments  are based on Pytorch Lightning\footnote{https://github.com/Lightning-AI/pytorch-lightning} and use torchvision's\footnote{https://pytorch.org/vision/stable/index.html} versions of the datasets.

The \texttt{MNIST}~\cite{lecun1998mnist} dataset consists of 60.000 train and 10.000 test samples. The \texttt{CIFAR10}~\cite{krizhevsky2009learning} and \texttt{CIFAR100}~\cite{krizhevsky2009learning} datasets both contains 50.000 train and 10.000 test samples. All models are trained from scratch. We calculate the accuracy as the mean of the percentage of correct batches across the test set.

% \begin{itemize}
%     \item indflydelse forwards / vs dropout quantisering 
%     \item brugen af parametre -> Skal vi bruge flere parametere for at modelere kapacitet i et ternary system?
%     \item Fungerer 1.58 bits some en regulariser (because it is coarse): I såfald kan mindre dropout på små netværk være en nødvendighed / fordel. Derudover skal L2 regularization også tones ned?
% \end{itemize}

% while weight decay on small language models has little effect for 16-bit and 1.58-bit, on image classification tasks with small models deadly on 16-bit, 1.58-bit much more robust
% 2 FIGURES: training accuracy 16-bit for different decay values, training accuracy 1.58-bit side by side with the same scale of y-axis

% while for small language models, we see some robustness as in bitnet 1.58 article, for small vision models, robustness to learning rate is not very pronounced compared to 16-bit; 
% 2 FIGURES for SLM: 16 and 1.58 side by side
% 2 FIGURES for image classification: 16 and 1.58 side by side

% less weight decay worse peformance for median, better performance for mean

\begin{figure}[htbp] 
  \centering
  \caption{Scaling behaviour of 16-bit and 1.58-bit (mean and medium) training for SLMs over 10 epochs (= 1{,}020 evaluations on test set.)}
  \begin{subfigure}[b]{0.48\textwidth}
    \centering
    \includegraphics[width=\textwidth]{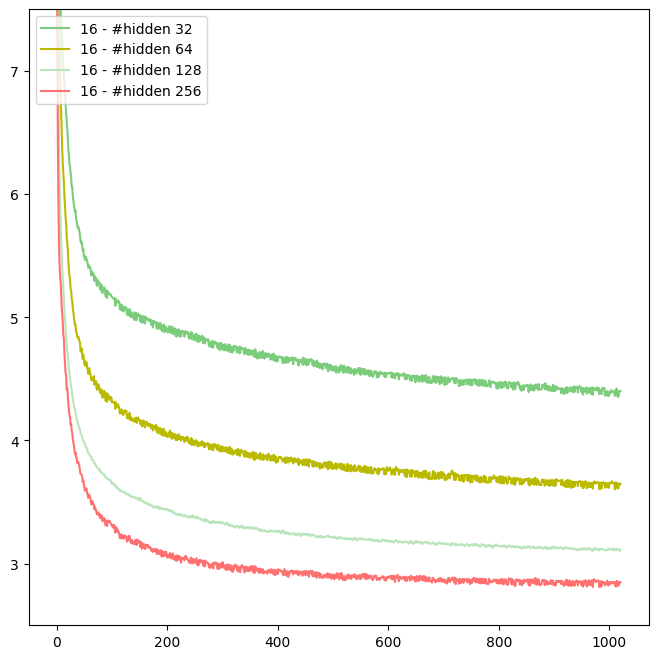}
    \caption{Scaling for 16 bit.}
    \label{fig:slm_scaling_16}
  \end{subfigure}
  \hfill
  \begin{subfigure}[b]{0.48\textwidth}
    \centering
    \includegraphics[width=\textwidth]{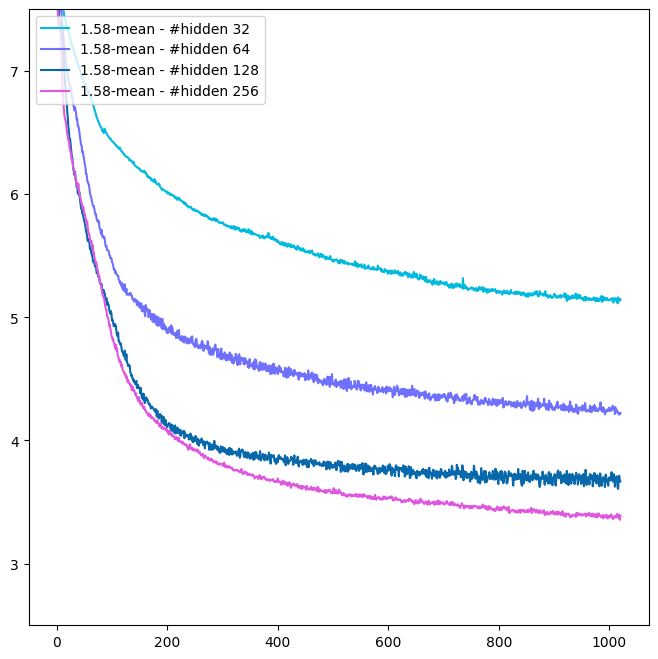}
    \caption{Scaling for 1.58 bit (mean).}
    \label{fig:slm_scaling_1.58}
  \end{subfigure}
  \hfill
  \begin{subfigure}[b]{0.48\textwidth}
    \centering
    \includegraphics[width=\textwidth]{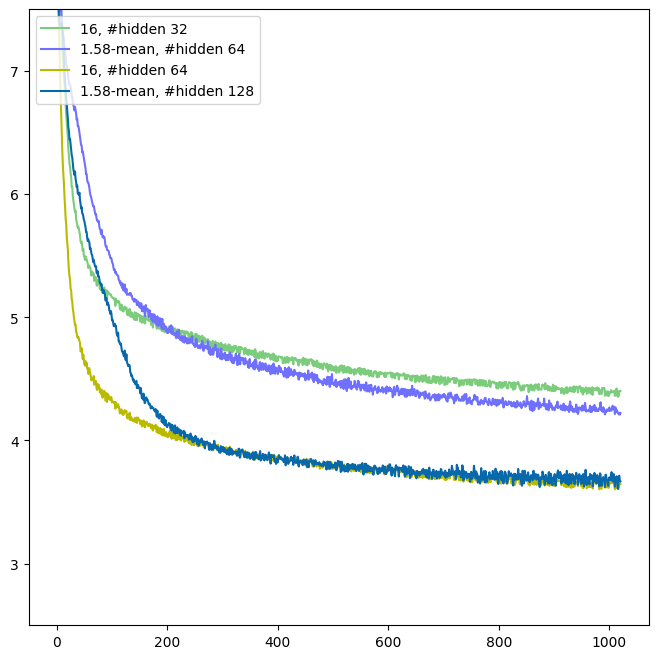}
    \caption{Scaling 16 vs 1.58 bit (mean).}
    \label{fig:slm_scaling_16_1.58}
  \end{subfigure}
  \hfill
  \begin{subfigure}[b]{0.48\textwidth}
    \centering
    \includegraphics[width=\textwidth]{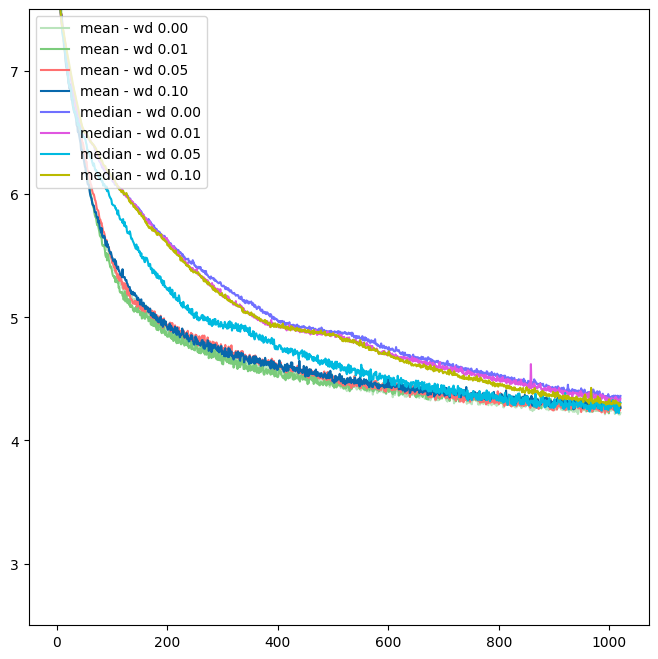}
    \caption{mean vs median for 64 hidden size.}
    \label{fig:slm_scaling_mean_median}
  \end{subfigure}
   \label{fig:slm_scaling}
\end{figure}

\section{Results}\label{sec:results}
In this section, we present a comparison of our \texttt{BitLinear} implementation with 16-bit floating point \texttt{torch.nn.Layer}, showing close-to-state-of-the-art performance on SLMs and better-than-state-of-the-art performance for vision models. We also perform ablation studies on the learning rate and weight decay hyperparameters, as well as the choice of mean vs median for the quantization of the weights.

\subsection{Small Language Models}\label{sec:results:slm}
The first experiment for SLMs is a scaling experiment, where we perform 16-bit and 1.58-bit training on all four model sizes. The second experiment is a hyperparameter tuning for the  learning rate and weight decay in a 12M SLM, with a fixed hidden size of 64. We show the results of the first and second experiment in Tables~\ref{tab:slm_scaling}~and~\ref{tab:slm_hyperparameters}, respectively.
% \peter{Describe also second experiment (LR and WD for fixed 64 hidden size), pointing to table 2.}

\begin{figure}[htbp] 
  \centering
  \caption{Hyperparameter tuning regarding weight decay and learning rate for SLMs over 10 epochs (= 1{,}020 evaluations on test set.)}
  \begin{subfigure}[b]{0.48\textwidth}
    \centering
    \includegraphics[width=\textwidth]{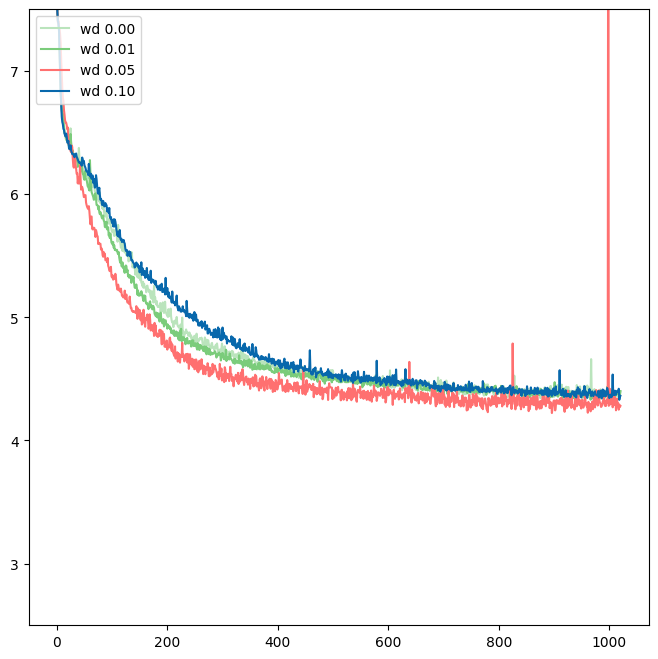}
    \caption{Weight decay for 1.58 bit (mean)}
    \label{fig:slm_wd_1.58_mean}
  \end{subfigure}
  \hfill
  \begin{subfigure}[b]{0.48\textwidth}
    \centering
    \includegraphics[width=\textwidth]{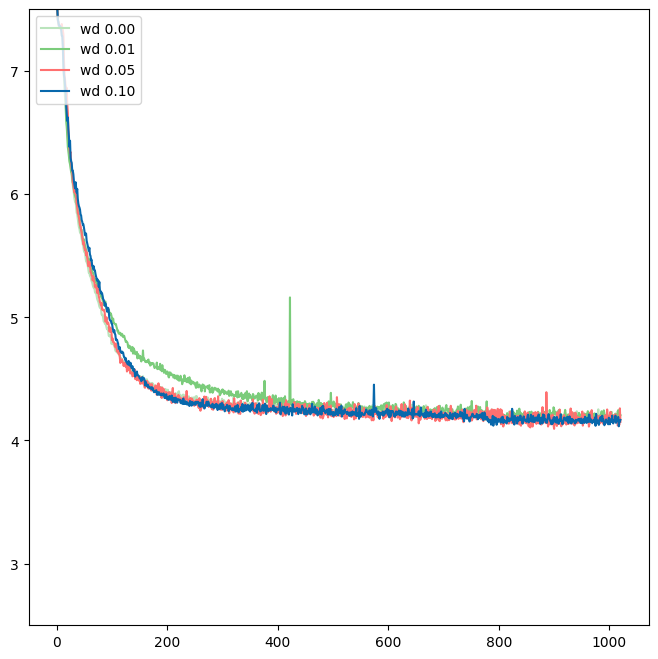}
    \caption{Weight decay for 1.58 bit (median)}
    \label{fig:slm_wd_1.58_median}
  \end{subfigure}
  \hfill
  \begin{subfigure}[b]{0.48\textwidth}
    \centering
    \includegraphics[width=\textwidth]{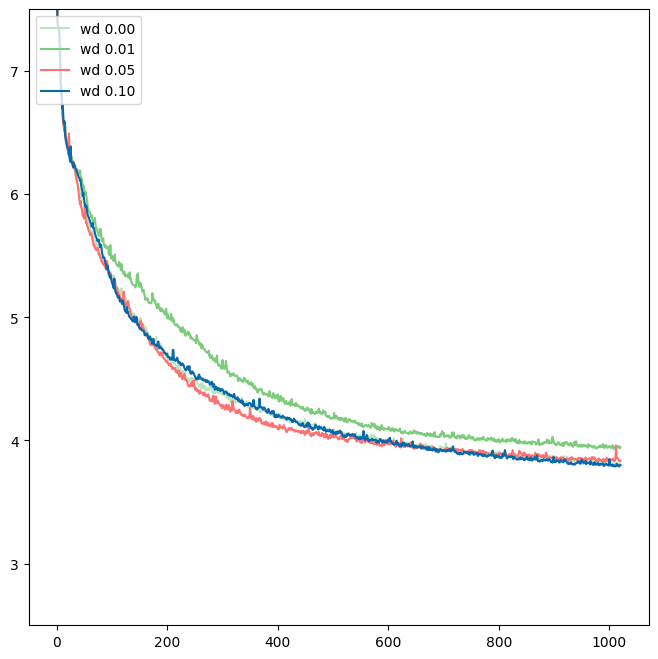}
    \caption{Weight decay for 16 bit}
    \label{fig:slm_wd_16}
  \end{subfigure}
  \hfill
  \begin{subfigure}[b]{0.48\textwidth}
    \centering
    \includegraphics[width=\textwidth]{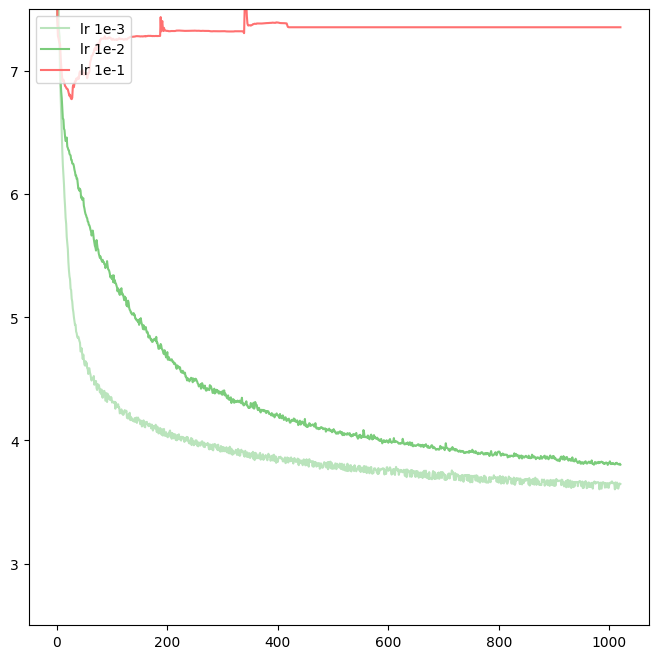}
    \caption{Learning rate for 16 bit}
    \label{fig:slm_lr_16}
  \end{subfigure}
  \hfill
  \begin{subfigure}[b]{0.48\textwidth}
    \centering
    \includegraphics[width=\textwidth]{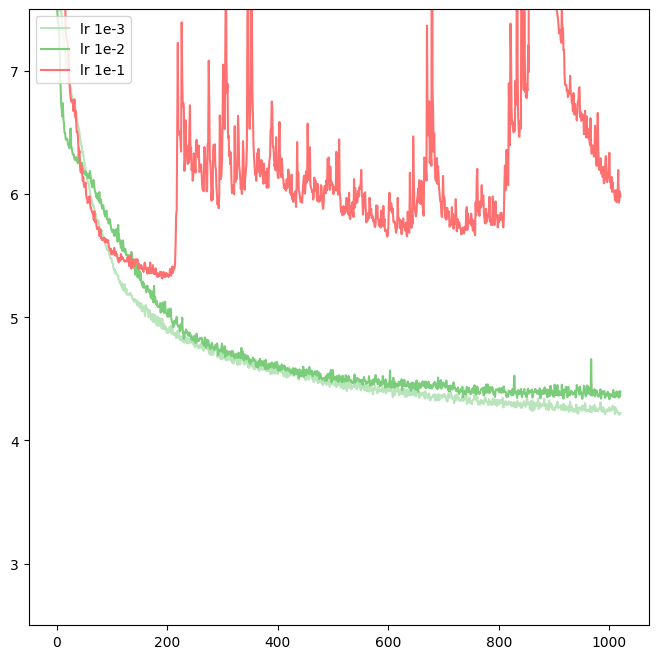}
    \caption{Learning rate for 1.58 bit (mean)}
   \label{fig:slm_lr_1.58_mean}
  \end{subfigure}
  \hfill
  \begin{subfigure}[b]{0.48\textwidth}
    \centering
    \includegraphics[width=\textwidth]{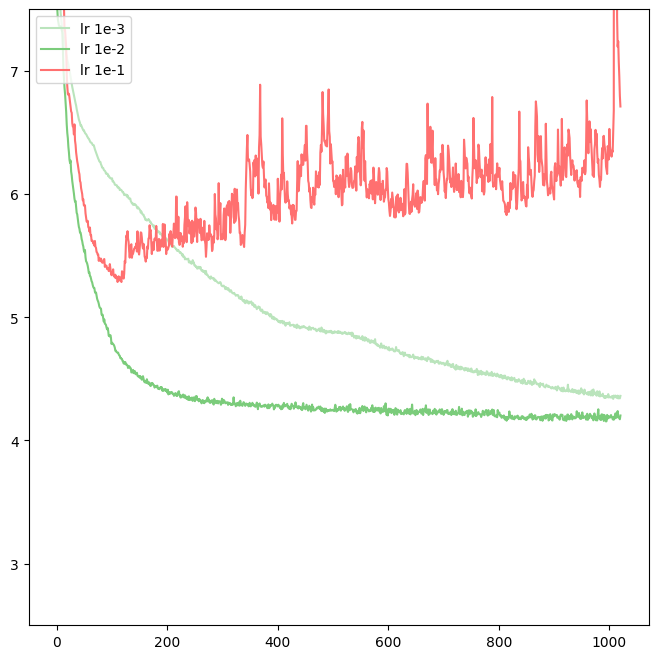}
    \caption{Learning rate for 1.58 bit (median)}
   \label{fig:slm_lr_1.58_median}
  \end{subfigure}
   \label{fig:slm_hyper}
\end{figure}

Both tables show the different configurations and perplexities after 10 epochs. For most configurations, the training has converged or is close to convergence at the end of the experiment. It is important to keep in mind that the reported perplexity is the exponentiation of the entropy, i.e., here the exponentiation of the loss defined via cross-entropy. Thus, minor changes in the loss result in quite discernible changes to the perplexity.

The first two columns give the hidden layer size and the number of parameters. The third column provides the bit-depth and implementation: ``16'' stands for 16-bit training, ``1.58-mean'' for our \texttt{BitLinear} implementation with 1.58 bits and AbsMean quantization of weights, and ``1.58-median'' for our \texttt{BitLinear} implementation with 1.58 bits and AbsMedian quantization of weights.

We show the results of this first experiment in Figure \ref{fig:slm_scaling}. Figure~\ref{fig:slm_scaling_16}) show that the 16-bit training scales exactly as expected when the number of hidden layers, and thus the models capacity, increases. We see in Figure \ref{fig:slm_scaling_1.58}, that 1.58-bit training follows the same trend, albeit with slightly lower performance. In Figure \ref{fig:slm_scaling_16_1.58}, we can visually compare the scaling between 16-bit training for models with 32 and 64 hidden sizes and 1.58-bit training for models with 64 and 128 hidden sizes. The observed perplexities suggest that the effective capacities of the models with 1.58-bit weights are around half that of the models with 16-bit weights, i.e., that hidden layers of approximately double size are needed for 1.58-bit models to reach performance comparable with the 16-bit counterparts. 
Figure \ref{fig:slm_scaling_mean_median} shows that the median generally converges slower than the mean over the employed weight decays. We discuss this fact in Section~\ref{sec:discussion}.  
% \peter{Say something about 2d), i.e., that median converges slower and why ...}

The fourth column shows the learning rate. For 16-bit training, we took a high but stable learning rate of 0.001 (1e-3). For 1.58-bit, we used the same or a larger learning rate of 0.01 (1e-2), as 1.58-bit training has been found to be more robust to higher learning rate in the context of LLMs\cite{ma2024era}. The fifth column shows the weight decay. We tried both a small but noticeable decay of 5\%, which is pretty prevalent in the pre-training and fine-tuning of LLMs, and no weight decay. Both Tables~\ref{tab:slm_scaling}~and~\ref{tab:slm_hyperparameters} hint that a weight decay of 5\% yields the best or similar performance compared to other values of weight decay. This is also visualized in Figures \ref{fig:slm_wd_1.58_mean}, \ref{fig:slm_wd_1.58_median}, and \ref{fig:slm_wd_16}, where trainings with a weight decay of 5\% are represented by a red line. 

% \peter{Summarize for LR and WD what we see here - 0.05 seems to be the sweet across 16, 1.58-mean, and 1.58-median.}

The sixth column provides the perplexity after 10 epochs. For nearly all configurations, after 10 epochs the training had converged. The best perplexities for 16-bit and 1.58-bit training are marked in bold, respectively. The seventh and last column shows the number of epochs, with total training length corresponding to 135M tokens per epoch, i.e., 1.35B tokens per 10-epoch experiment.

\begin{table}[htb]
  \centering
  % \small
  \scriptsize 
  \caption{\small{Language modelling benchmarks at different scales.}}
  \setlength{\tabcolsep}{3pt}  % Increase spacing to 6pt (adjust as needed)
  \begin{tabular}{lcccccc}
  \midrule
  \textbf{\#Hidden} & \textbf{\#Params} & \textbf{Bits} & \textbf{Learning Rate} & \textbf{Weight Decay} & \textbf{Perplexity} & \textbf{Epochs}\\
  \midrule
  \multirow{12}{*}{32} & \multirow{12}{*}{6M} & \multirow{2}{*}{16} & \multirow{2}{*}{0.001} & 0.00 & \textbf{77.8} & 10\\
  &&&& 0.05 & 81.0 & 10\\
  \cmidrule(l){3-7}
  && \multirow{4}{*}{1.58-mean} & \multirow{2}{*}{0.001} & 0.00 & 166.2 & 10\\
  &&&& 0.05 & 164.9 & 10\\
  \addlinespace
  &&& \multirow{2}{*}{0.01} & 0.00 & 130.1 & 10\\
  &&&& 0.05 & 134.4 & 10\\
  \cdashlinel{3-7}
  && \multirow{4}{*}{1.58-median} & \multirow{2}{*}{0.001} & 0.00 & 183.6 & 10\\
  &&&& 0.05 & 183.8 & 10\\
  \addlinespace
  &&& \multirow{2}{*}{0.01} & 0.00 & \textbf{116.6} & 10\\
  &&&& 0.05 & 118.0 & 10\\
  \midrule
  \multirow{12}{*}{64} & \multirow{12}{*}{12M} & \multirow{2}{*}{16} & \multirow{2}{*}{0.001} & 0.00 & \textbf{36.7} & 10\\
  &&&& 0.05 & 37.5 & 10\\ 
  \cmidrule(l){3-7}
  && \multirow{4}{*}{1.58-mean} & \multirow{2}{*}{0.001} & 0.00 & 67.4 & 10\\ 
  &&&& 0.05 & 68.2 & 10\\ 
  \addlinespace
  &&& \multirow{2}{*}{0.01} & 0.00 & 76.3 & 10\\ 
  &&&& 0.05 & 68.2 & 10\\ 
  \cdashlinel{3-7}
  && \multirow{4}{*}{1.58-median} & \multirow{2}{*}{0.001} & 0.00 & 76.5 & 10\\ 
  &&&& 0.05 & 68.1 & 10\\ 
  \addlinespace
  &&& \multirow{2}{*}{0.01} & 0.00 & 61.1 & 10\\
  &&&& 0.05 & \textbf{60.0} & 10\\
  \midrule
  \multirow{12}{*}{128} & \multirow{12}{*}{24M} & \multirow{2}{*}{16} & \multirow{2}{*}{0.001} & 0.00 & 22.3 & 10\\
  &&&& 0.05 & \textbf{21.4} & 10\\
  \cmidrule(l){3-7}
  && \multirow{4}{*}{1.58-mean} & \multirow{2}{*}{0.001} & 0.00 & 36.8 & 10\\ 
  &&&& 0.05 & \textbf{36.3} & 10\\ 
  \addlinespace
  &&& \multirow{2}{*}{0.01} & 0.00 & 61.6 & 10\\ 
  &&&& 0.05 & 71.0 & 10\\ 
  \cdashlinel{3-7}
  && \multirow{4}{*}{1.58-median} & \multirow{2}{*}{0.001} & 0.00 & 39.8 & 10\\ 
  &&&& 0.05 & 37.5 & 10\\ 
  \addlinespace
  &&& \multirow{2}{*}{0.01} & 0.00 & 42.3 & 10\\
  &&&& 0.05 & 38.4 & 10\\
  \midrule
  \multirow{12}{*}{256} & \multirow{12}{*}{48M} & \multirow{2}{*}{16} & \multirow{2}{*}{0.001} & 0.00 & \textbf{16.6} & 10\\
  &&&& 0.05 & 16.7 & 10\\
  \cmidrule(l){3-7}
  && \multirow{4}{*}{1.58-mean} & \multirow{2}{*}{0.001} & 0.00 & 28.7 & 10\\ 
  &&&& 0.05 & 27.1 & 10\\ 
  \addlinespace
  &&& \multirow{2}{*}{0.01} & 0.00 & 77.7 & 10\\ 
  &&&& 0.05 & 65.6 & 10\\ 
  \cdashlinel{3-7}
  && \multirow{4}{*}{1.58-median} & \multirow{2}{*}{0.001} & 0.00 & \textbf{26.8} & 10\\ 
  &&&& 0.05 & 27.5 & 10\\ 
  \addlinespace
  &&& \multirow{2}{*}{0.01} & 0.00 & 65.1 & 10\\
  &&&& 0.05 & 63.8 & 10\\
  \midrule
  \end{tabular}
  \label{tab:slm_scaling}
  \vspace{-2ex}
\end{table}

\begin{table}
  \centering
  % \small
  \scriptsize 
  \caption{\small{Hyperparameter tuning for a 12M SLM.}}
  \setlength{\tabcolsep}{3pt}  % Increase spacing to 6pt (adjust as needed)
  \begin{tabular}{lcccccc}
  \midrule
  \textbf{\#Hidden} & \textbf{\#Params} & \textbf{Bits} & \textbf{Learning Rate} & \textbf{Weight Decay} & \textbf{Perplexity} & \textbf{Epochs}\\
  \midrule
  \multirow{40}{*}{64} & \multirow{40}{*}{12M} & \multirow{12}{*}{16} & \multirow{4}{*}{0.001} & 0.00 & \textbf{36.7} & 10\\
  &&&& 0.01 & 37.7 & 10\\ 
  &&&& 0.05 & 37.5 & 10\\
  &&&& 0.10 & 36.8 & 10\\ 
  \addlinespace
  &&& \multirow{4}{*}{0.01} & 0.00 & 44.8 & 10\\ 
  &&&& 0.01 & 51.0 & 10\\ 
  &&&& 0.05 & 45.9 & 10\\ 
  &&&& 0.10 & 44.2 & 10\\ 
  \addlinespace
  &&& \multirow{4}{*}{0.1} & 0.00 & 871.5 & 10\\ 
  &&&& 0.01 & 938.9 & 10\\ 
  &&&& 0.05 & 191.3 & 10\\ 
  &&&& 0.10 & 65.3 & 10\\ 
  \cmidrule(l){3-7}
  && \multirow{12}{*}{mean} & \multirow{4}{*}{0.001} & 0.00 & 67.4 & 10\\ 
  &&&& 0.01 & 68.1 & 10\\ 
  &&&& 0.05 & 68.2 & 10\\ 
  &&&& 0.10 & 69.6 & 10\\ 
  \addlinespace
  &&& \multirow{4}{*}{0.01} & 0.00 & 76.3 & 10\\ 
  &&&& 0.01 & 76.0 & 10\\ 
  &&&& 0.05 & 68.2 & 10\\ 
  &&&& 0.10 & 76.0 & 10\\ 
  \addlinespace
  &&& \multirow{4}{*}{0.1} & 0.00 & 203.3 & 10\\ 
  &&&& 0.01 & 240.4 & 10\\ 
  &&&& 0.05 & 173.5 & 10\\ 
  &&&& 0.10 & 204.5 & 10\\
  \cdashlinel{3-7}
  && \multirow{12}{*}{median} & \multirow{4}{*}{0.001} & 0.00 & 76.5 & 10\\ 
  &&&& 0.01 & 74.1 & 10\\ 
  &&&& 0.05 & 68.1 & 10\\ 
  &&&& 0.10 & 72.2 & 10\\ 
  \addlinespace
  &&& \multirow{4}{*}{0.01} & 0.00 & 61.1 & 10\\
  &&&& 0.01 & 63.5 & 10\\
  &&&& 0.05 & \textbf{60.0} & 10\\
  &&&& 0.10 & 61.3 & 10\\
  \addlinespace
  &&& \multirow{4}{*}{0.1} & 0.00 & 197.6 & 10\\
  &&&& 0.01 & 223.2 & 10\\
  &&&& 0.05 & 154.0 & 10\\
  &&&& 0.10 & 157.9 & 10\\
  \midrule
  \midrule
  \end{tabular}
  \label{tab:slm_hyperparameters}
  \vspace{-2ex}
\end{table}

\subsection{Small Vision Models}\label{sec:results:svm}
% setup

\begin{figure}[htbp] 
  \centering
  \caption{The effect of the learning rate (LR) on the training robustness for \texttt{CIFAR100} over 10 epochs.}
  \begin{subfigure}[b]{0.48\textwidth}
    \centering
    \includegraphics[width=\textwidth]{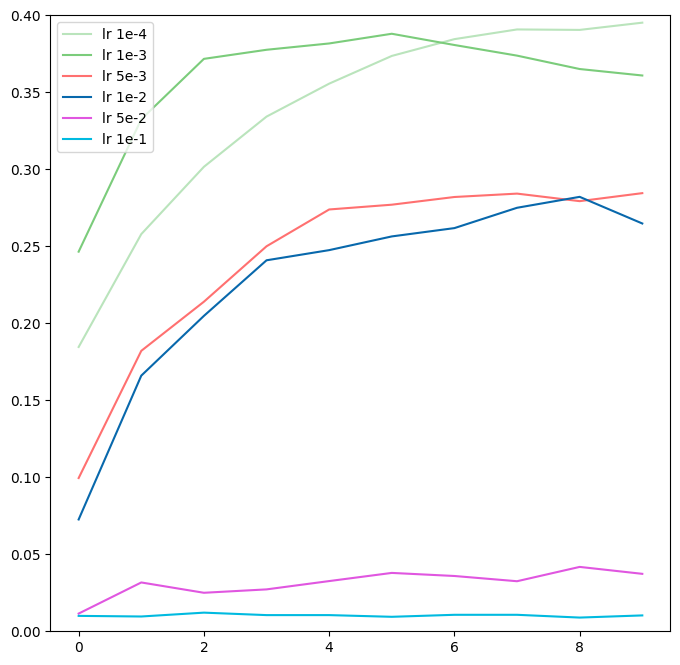}
    \caption{1.58-Bit (mean).}
    \label{fig:lr_robustness_cifar100_158bit}
  \end{subfigure}
  \hfill
  \begin{subfigure}[b]{0.48\textwidth}
    \centering
    \includegraphics[width=\textwidth]{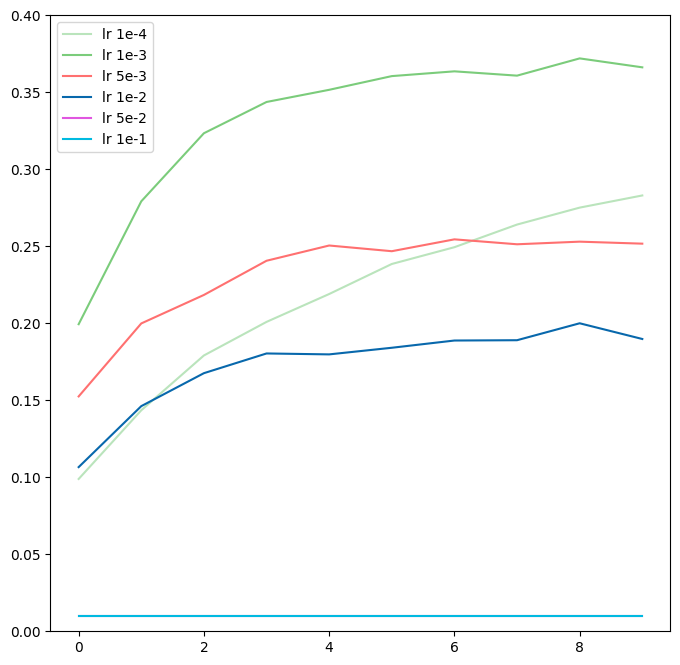}
    \caption{16-bit.}
   \label{fig:lr_robustness_cifar100_16bit}
   \label{fig:lr_robustness_cifar100}
  \end{subfigure}
\end{figure}

% high small learning rate
% \peter{Don't start a sentence with a cite ... The implementation of BitNet b1.58 cite tips code adopts ... also next sentence!}
The implementation of Bitnet b1.58 \cite{158bit_tips_code} adopts from \cite{ma2024era} the strategy of employing significantly higher learning rates, arguing that this is crucial for optimising the 1.58-bit weights. They also state, that this does not carry to the full 16-bit precision, suggesting this might be because of prior fine-tuning. 
We show in the graphs presented in Figure \ref{fig:lr_robustness_cifar100} that larger learning rates are sub-optimal for both 1.58- and 16-bit weights in small classification models, despite being trained from scratch. We use the mean-based benchmark for comparability, but observe similar results for the median-based counterpart. For 1.58 bits, we see in Figure~\ref{fig:lr_robustness_cifar100_158bit} that performance gradually declines as the learning grows from 0.0001 to 0.1, with the smallest learning rate providing the best performance. We observe a similar trend in Figure \ref{fig:lr_robustness_cifar100_16bit}, where learning with a rate of 0.05 or above distorts the training, preventing the network from learning at all, as evident in the evaluation.

\begin{figure}[htbp] 
  \centering
  \caption{The effect of weight decay (WD) on the training robustness for \texttt{CIFAR100} over 10 epochs.}
  \begin{subfigure}[b]{0.48\textwidth}
    \centering
    \includegraphics[width=\textwidth]{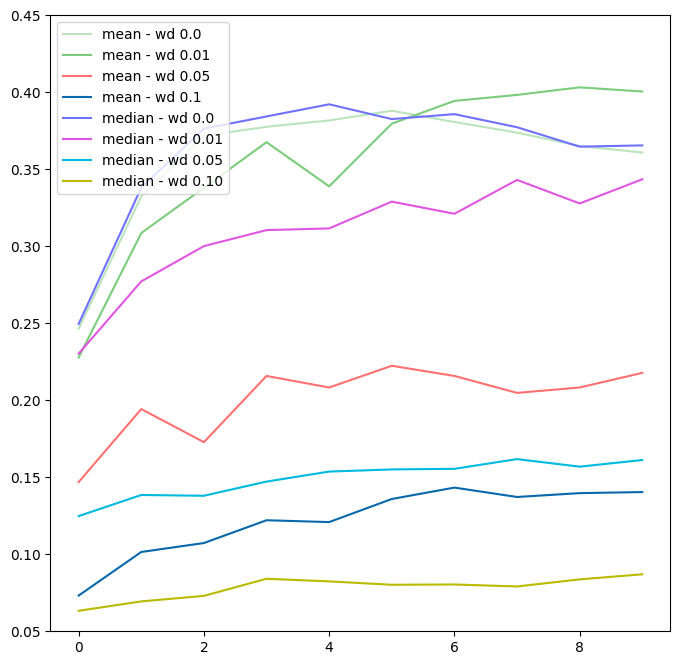}
    \caption{Weight decay for learning rate 0.001.}
    \label{fig:lr_robustness_to_weight_decay_cifar100_0.001}
  \end{subfigure}
  \hfill
  \begin{subfigure}[b]{0.48\textwidth}
    \centering
    \includegraphics[width=\textwidth]{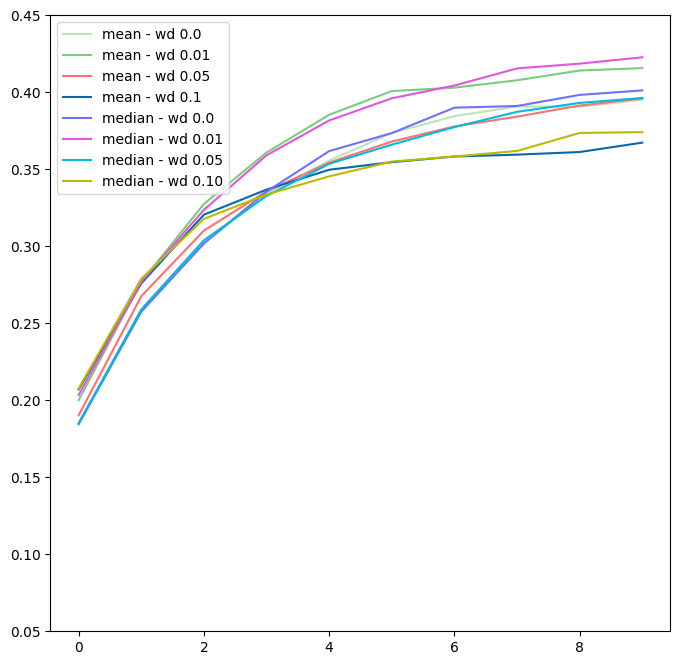}
    \caption{Weight decay for learning rate 0.0001.}
   \label{fig:lr_robustness_to_weight_decay_cifar100_0.0001}
  \end{subfigure}
   \label{fig:lr_robustness_to_weight_decay_cifar100}
\end{figure}

In Figure~\ref{fig:lr_robustness_to_weight_decay_cifar100} we document the impact of weight decays of 0\%, 1\%, 5\%, and 10\% across the two learning rates 0.001 and 0.0001. Figure~\ref{fig:lr_robustness_to_weight_decay_cifar100_0.001} showcases the effect of weight decay when using higher learning rates, whereas in Figure \ref{fig:lr_robustness_to_weight_decay_cifar100_0.0001}, we see more continuity over the first epochs and that a training with a weight decay of 1\% appears superior.

% mean vs median
As described in Section~\ref{sec:method}, we conducted experiments with both AbsMean and with AbsMedian quantization, all of which are shown in Table~\ref{tab:vision_baselines}. 
The mean-based quantization is superior on \texttt{MNIST} and \texttt{CIFAR10}, with $0.15$ and $1.22$ difference in percentage points test accuracy, respectively. On \texttt{CIFAR100} the median-based quantization is superior with a percentage point difference of $0.7$. From our experiments in Table~\ref{tab:vision_baselines}, no clear conclusion can be drawn as to which is preferable in general. Similarly, in Figure~\ref{fig:lr_robustness_to_weight_decay_cifar100_0.0001}, we do not see a clear distinction between the two in neither the evolving performance nor the resulting one.
Therefore, we propose the choice of AbsMean vs AbsMedian quantization for the weights as a hyperparameter for 1.58-bit training.
% \peter{the previous sentence does not make sense ...}

% weight decay
We investigate the effect of weight decay on the 1.58-bit networks and compare to 16-bit benchmarks in Table~\ref{tab:vision_baselines}. We employed weight decays of 0\%, 1\%, 5\%, and 10\%. For both \texttt{CIFAR10} and \texttt{CIFAR100}, the 1.58-bit models are significantly more robust compared to their 16-bit counterparts, which become so unstable the training is distorted and evaluates to a test accuracy of $1.00$. This might be because of the coarse training-scheme inherently associated with 1.58-bit quantization, increasing the robustness of the training against further regularization through, for example, weight decay. For \texttt{MNIST} we see that, while weight decay yields a decrease in accuracy, is does not prevent the models from learning. Similarly, we see that the 1.58-bit training present more stability in performance when weight decay is employed. 
% \peter{rest of sentence missing}

% We show that such higher learning rates are sub-optimal for classification tasks on small networks, evidently shown on experiments in Table \ref{tab:vision_baselines}. 
% We show in FIGURES, that these small networks yields more instability. We further show in FIGURES that weight decay is not aiding training, introducing instability in the training. However, this might not have something to do with the 1-Bit weights, but more something to do with the small networks getting suffocated. 

% \begin{itemize}
%     \item The hard weight decay make sense on MNIST as for each of the number 0-9, we want a very strong signal on the TP classification, we blur this out by enforcing weight decay. - Maybe this carries to classification in general where overfitting can be a problem?
%     \item Above seems to carry with a lot more effect to CIFAR100 
%     \item seems above fact follows conventional small networks. Not intersting in itself, however, it actually confirms the capbalities of BitLinear. 
% \end{itemize}

\begin{table}[htb]
    \centering
    % \small
    \scriptsize 
    \caption{\small{Supervised benchmarks on vision classification datasets.}}
    \setlength{\tabcolsep}{3pt}  % Increase spacing to 6pt (adjust as needed)
    \begin{tabular}{lcccccc}
    \midrule
    \textbf{Dataset} &  \textbf{\#Params} &  \textbf{Bits} & \textbf{Learning Rate} & \textbf{Weight Decay} & \textbf{Test Accuracy} & \textbf{Epochs}\\
    \midrule
    \multirow{25}{*}{MNIST} & \multirow{25}{*}{100K} 
                           & \multirow{5}{*}{16} & 0.0001 & 0.00 & 92.29 & 10 \\
                           \addlinespace
                           &&& \multirow{4}{*}{0.001} & 0.00 & \textbf{96.93} & 10 \\
                           &&&& 0.01 & 93.35 & 10 \\
                           &&&& 0.05 & 77.06 & 10 \\
                           % &&&& 0.10 & 11.35 & 10 \\
                           \cmidrule(l){3-7}
                  
                           && \multirow{7}{*}{1.58-mean} & 0.0001 & 0.00 & 95.63 & 10 \\
                           &&& 0.0001 & 0.01 & \textbf{96.08} & 10 \\
                           \addlinespace
                           &&& \multirow{3}{*}{0.001} & 0.00 & 96.01 & 10 \\
                           &&&& 0.01 & 93.11 & 10 \\
                           &&&&  0.05 & 86.57 & 10 \\
                           \addlinespace
                           &&& 0.01 & 0.00 & 94.59 & 10 \\
                           &&& 0.05 & 0.00 & 93.80 & 10 \\
                           &&& 0.10 & 0.00 & 93.15 & 10 \\
                           
                           \cdashlinel{3-7}
                           
                           && \multirow{7}{*}{1.58-median} & 0.0001 & 0.00 & 94.14 & 10 \\
                           &&& 0.0001 & 0.01 & 95.93 & 10 \\
                           % &&& 0.0001 & 0.05 & 95.50 & 10 \\
                           \addlinespace
                           &&& \multirow{3}{*}{0.001} & 0.00 & 95.80 & 10 \\
                           &&&& 0.01 & 91.27 & 10 \\
                           &&&& 0.05 & 89.15 & 10 \\
                           \addlinespace
                           &&& 0.01 & 0.00 & 93.03 & 10 \\
                           &&& 0.05 & 0.00 & 86.61 & 10 \\
                           &&& 0.10 & 0.00 & 52.35 & 10 \\

    \midrule
    \multirow{25}{*}{CIFAR10} & \multirow{25}{*}{2.1M} 
                            & \multirow{5}{*}{16} & 0.0001 & 0.00 & 60.86 & 10 \\
                            \addlinespace
                            &&& \multirow{4}{*}{0.001} & 0.00 & \textbf{70.06} & 10 \\
                            &&&& 0.01 & 58.32 & 10 \\
                            &&&& 0.05 & 10.0 & 10 \\
                            &&&& 0.10 & 10.0 & 10 \\
                            \cmidrule(l){3-7}
                 
                            && \multirow{10}{*}{1.58-mean}  & \multirow{3}{*}{0.0001} & 0.00 & 68.94 & 10 \\
                            &&&& 0.01 & 69.1 & 10 \\
                            &&&& 0.05 & \textbf{71.47} & 10 \\
                            % &&& 0.0001 & 0.10 & 70.26 & 10 \\
                            \addlinespace
                            
                            &&& \multirow{3}{*}{0.001} & 0.00 & 70.35 & 10 \\
                            &&&& 0.01 & 69.08 & 10 \\
                            &&&& 0.05 & 58.04 & 10 \\
                            \addlinespace
                            &&& 0.01 & 0.00 & 63.92 & 10 \\
                            &&& 0.05 & 0.00 & 25.01 & 10 \\
                            &&& 0.10 & 0.00 & 23.05 & 10 \\
  
                            \cdashlinel{3-7}
                            && \multirow{10}{*}{1.58-median} & \multirow{3}{*}{0.0001} & 0.00 & 69.08 & 10 \\
                            &&&& 0.01 & 69.55 & 10 \\
                            &&&& 0.05 & 70.25 & 10 \\
                            % &&& 0.0001 & 0.10 & 69.83 & 10 \\
                            
                            \addlinespace
                            &&& \multirow{3}{*}{0.001} & 0.00 & 71.21 & 10 \\
                            &&&& 0.01 & 69.80 & 10 \\
                            &&&& 0.05 & 60.61 & 10 \\
                            \addlinespace
                            &&& 0.01 & 0.00 & 65.80 & 10 \\
                            &&& 0.05 & 0.00 & 54.77 & 10 \\
                            &&& 0.10 & 0.00 & 49.48 & 10 \\
                           
    \midrule
    \end{tabular}
    \label{tab:vision_baselines}
    \vspace{-2ex}
\end{table}
\begin{table}[htb]
    \centering
    % \small
    \scriptsize 
    \setlength{\tabcolsep}{3pt}  % Increase spacing to 6pt (adjust as needed)
    \begin{tabular}{lcccccc}
    \midrule
    \textbf{Dataset} &  \textbf{\#Params} &  \textbf{Bits} & \textbf{Learning Rate} & \textbf{Weight Decay} & \textbf{Test Accuracy} & \textbf{Epochs}\\
    \midrule
     \multirow{25}{*}{CIFAR100} & \multirow{25}{*}{2.2M} 
                               & \multirow{5}{*}{16} & 0.0001 & 0.00 & 28.30 & 10 \\
                            \addlinespace
                               &&& \multirow{4}{*}{0.001} & 0.00 & \textbf{36.62} & 10 \\
                               
                               &&&& 0.01 & 17.97 & 10 \\
                               &&&& 0.05 & 1.0 & 10 \\
                               % &&&& 0.10 & 1.0 & 10 \\
                            \addlinespace
                               % &&& 0.005 & 0.00 & 25.17 & 10 \\
                               % &&& 0.01 & 0.00 & 1.0 & 10 \\
                               % && None & 0.05 & 0 & 1.0 & 10 \\
                               
                               \cmidrule(l){3-7}
                               
                               % \cmidrule(l){3-7}
                               && \multirow{6}{*}{1.58-mean} & \multirow{3}{*}{0.0001} & 0.00 & 39.52 & 10 \\
                               &&&& 0.01 & 41.57 & 10 \\
                               &&&& 0.05 & 39.60 & 10 \\
                            \addlinespace
                               &&&  \multirow{3}{*}{0.001} & 0.00 & 36.09 & 10 \\ % wd_0.01: 40.05
                               &&&& 0.01 & 40.05 & 10 \\
                               &&&& 0.05 & 21.78 & 10 \\
                               \addlinespace
                               &&& 0.01 & 0.00 & 26.48 & 10 \\
                               &&& 0.05 & 0.00 & 3.73 & 10 \\
                               &&& 0.10 & 0.00 & 1.03 & 10 \\
                            
                               \cdashlinel{3-7}
                               && \multirow{6}{*}{1.58-median} & \multirow{3}{*}{0.0001} & 0.00 & 40.12 & 10 \\
                               &&&& 0.01 & \textbf{42.27} & 10 \\
                               &&&& 0.05 & 39.63 & 10 \\
                            \addlinespace
                               &&& \multirow{3}{*}{0.001} & 0.00 & 36.55 & 10 \\
                               &&&& 0.01 & 34.35 & 10 \\
                               &&&& 0.05 & 16.12 & 10 \\
                               % &&& 0.001 & 0.00 & 36.55 & 10 \\
                               \addlinespace
                               &&& 0.01 & 0.00 & 30.06 & 10 \\
                               &&& 0.05 & 0.00 & 5.53 & 10 \\
                               &&& 0.10 & 0.00 & 1.93 & 10 \\
                               
    \midrule
    \end{tabular}
    \vspace{-2ex}
  \end{table}

% \jacob{PLOT: Weight decay for the learning rate 0.001, for 16-Bit, 1.58bit mean and 1.58bit meadian}

% \jacob{PLOT: 1.58_cifar_median_vs_median_lr_0.0001.png. Showing the weight decay. 
%     - More important later in the training stage than in the beginning
%     }

% \begin{figure}[htbp] 
%   \centering
%   \caption{The learning rate's (LR) effect on training robustness for small vision models on \texttt{CIFAR100} over 10 epochs.}
%   \begin{subfigure}[b]{0.48\textwidth}
%     \centering
%     \includegraphics[width=\textwidth]{plots/cifar_100_158_bit.png}
%     \caption{1.58-Bit (mean)}
%     \label{fig:lr_robustness_cifar100_158bit}
%   \end{subfigure}
%   \hfill
%   \begin{subfigure}[b]{0.48\textwidth}
%     \centering
%     \includegraphics[width=\textwidth]{plots/cifar_100_16_bit.png}
%     \caption{16-bit}
%    \label{fig:lr_robustness_cifar100_16bit}
%   \end{subfigure}
%    \label{fig:lr_robustness_cifar100}
% \end{figure}

\section{Discussion}\label{sec:discussion}
% main take away: 1.58 bit also works for small models, sometimes with a penalty (small language models), sometimes overreaching 16-bit performance (image classification)
% NO FIGURES, JUST TABLES

% scaling by number of parameters works as expected for 1.58-bit, trailing the performance by approximately a binary order of magnitude, i.e., 2 the number of params for 16-bit are needed for similar performance with 1.58-bit; trade-off between number of params and bits per parameter, for larger models less pronounced as larger models have significant number of redundant/unused params (reference to the right article!)

% sometimes mean is best, sometimes median - which quantization to use to -1, 0, 1 is an open question for both small language and vision models (and probably also for large models); new hyperparameter
% NO FIGURES, JUST TABLES - SHORT DESCRIPTION

% \peter{jacob - come back to scaling of SLMs vs LLMs and refer to literature on unused parameters in LLMs!}
%
% should we write state-of-the-art performance?
Our results demonstrate that 1.58-bit training provides competitive performance on both small language and vision models. We hope our work encourages the community to work on 1.58-bit based architectures to facilitate efficient and fast inference independent of model scale. Overall, this enables both more environmentally friendly inference for many applications, and the deployment of deep neural networks in various low-resource uses-cases, with the potential of increased energy efficiency through multiplication-free kernels and even specialised hardware.

% scaling
As reported in Section \ref{sec:results}, using our approach to 1.58-bit training generally yields a small performance penalty in SLMs. In the small vision models, we see 1.58-bit outperforming the CNN-based networks for the \texttt{CIFAR10} and \texttt{CIFAR100} datasets, while being within a percentage point difference of $0.85$ for a sequential-based network trained on \texttt{MNIST}. This difference between training on text and image data is not entirely unexpected due to the nature of complexity difference in the SLM-data and the simpler vision-classification datasets. While 1.58-bit training still relies on the full precision 16-bit weights as shadow weights for the quantization in computing the 1.58-bit weights, we are reducing the capacity of each weight in the linear layers and, hence, of the overall network. 

This implies that, in some case, there might be a need for creating networks relying on an increased number of parameters to re-introduce some capacity. Evidently in our SLM-results shown in Figure \ref{fig:slm_scaling_16_1.58}, we see the need to utilize hidden layers of size 64 in the 1.58-mean case to gain the same performance as the 16-bit on using hidden layers of size 32. This is also holds for hidden layer sizes of 128 for 1.58 bits and corresponding 64 for 16-bit. Prior works have shown that LLMs do not utilize all parameters effectively~\cite{ashkboos2024slicegpt} and even consist of redundant layers~\cite{men2024shortgpt}. Therefore, we would expect the need for for this to decrease as model-size grows, i.e., we would expect 1.58-bit, to work well in networks from a certain size without increasing the number of parameters. This is in line with prior work \cite{ma2024era}. 

The need for increasing parameters seems to depend on the complexity of the downstream task, evident from our results on SLMs in Section~\ref{sec:results:svm} compared to our results on small vision models in Section~\ref{sec:results:svm}, where the same architecture (with an adjusted size of prediction head) outperforms full precision 16-bit models on both \texttt{CIFAR10} and \texttt{CIFAR100}. For SLMs, we do not consider the increased size of hidden layers to be an obstruction for profiting from 1.58-bit architectures, as the models can still be expected to run significantly more efficient in inference settings when implemented using custom kernels.

% mean vs median
In Sections \ref{sec:results:slm} and \ref{sec:results:svm} we conducted extensive experiments employing either one of the two quantization schemes: ``1.58-mean'' or ``1.58-median''. Changing between the two changes the factor with which 16-bit weights are scaled before rounding to integers in the quantization process. The median will, in some cases, be resilient to weight-updates, allowing higher variance without notable effect on the scaling factor. The mean will be more directly affected, particularly by large weight changes of few weights. Therefore, one provides flexibility of weights-updates whereas the other provides more constrained feedback affecting the gradient-magnitude on the shadow weights. 

 % Evident for both language and vision models in Tables \ref{tab:slm_scaling}, \ref{tab:slm_hyperparameters} and \ref{tab:vision_baselines}, one is not distinctive better than the other and therefore proposed as a hyper-parameter. The outcome does not seem to be related to model size nor hyper-parameters. Therefore, we probably would see the same for larger language models. Future research could try to shade light on this mechanism.

% \ref{fig:slm_lr}
% learning rate and robustness
% we see this exaclty in the plots, that we need the next over #params to achieve that 
% \jacob{We should explicitly mention the scaling-law/relation we see in the plots.}
% \jacob{Peter mentioned a specific article talking about redundancy or non-contributing parameters?}

In Figures \ref{fig:slm_lr_16}, \ref{fig:slm_lr_1.58_mean}, and \ref{fig:slm_lr_1.58_median}, we report the robustness in SLMs over different learning rates across 16-bit and both mean and median 1.58-bit schemes. Contrary, from what is reported on the 1.58-bit LLMs in \cite{158bit_tips_code}, the large learning rate $0.01$ (1e-1) produces instability to such a degree that it distorts the training, effectively rendering it unable to optimize the performance of the network. This also happens for 16 bits even though we are training from scratch. The fact that we are training small networks might explain this behaviour.  
The learning rate of $0.01$ (1e-2) in Figure \ref{fig:slm_lr_1.58_median} shows the effectiveness of the median quantization, as it converges faster than the mean quantization proposed in \cite{wang2023bitnet,ma2024era} and actually yields a convergent process similar to the one for 16 bits. This supports our claims of the behavior explained above, i.e., that the flexibility allowed in median quantization can aid faster convergence in some situations. Interestingly, we see that employing median quantization yields a significant difference in convergence when using a learning rate of $0.01$ and $0.001$ (1e-3), contrary to the same learning rates with mean quantization as displayed in Figure \ref{fig:slm_lr_1.58_mean}, making the network more sensitive to the learning rate. 
% This could mean, that there exist a relationship between the learning-rate the Measure-scheme employed and the values of the weight matrix (and variance of it), for future research to map. 
SMLs exhibt some but not the same level of learning-rate robustness as LLMs~\cite{ma2024era} when being trained using larger learning rates.

% - robustness between b1.58 and 16-bit
% - 64 not very good for 16 bits, better for mean, even better for median. 
% - the reason why a high learning-rate
% - 1.58 find out that, 16-bit does not benefit from high learning rates -> due to finetuning. We train from scratch, does not see the same effect, as on larger models. 
% - arugment that median provides some very benfitcial results.

% \subsubsection{Future Work}
% \begin{itemize}
%     \item ...
% \end{itemize}
\section{Conclusion}
In this paper, we introduced a variant of the BitNet b1.58-bit precision quantization-aware training demonstrating state-of-the-art performance on core downstream tasks for SLMs and vision classification models. To the best of our knowledge, this is the first work studying the characteristics and behaviour of the particular 1.58-bit quantization approach from \cite{wang2023bitnet,ma2024era} on small networks. The investigations provided in this work underline the potential of employing 1.58 bits more generally in small networks, mitigating prior arguments that these weight-resolutions only exhibits potential on large networks with billions of parameters. This opens up for the efficient deployment of SLMs and small vision models, particularly in low-resource use-cases. We encourage future work to investigate 1.58-bit quantization-aware training on other networks such as object-detection networks in the vision domain and language models with encoders, investigating the degree to which our conclusions hold for such types of networks.

% scaling + scaling law
Our results suggest a scaling law for small SLMs, with a 1.58-bit network needing approximately the double size of hidden layers to achieve performance comparable to 16-bit versions.
% learning rate
The learning rate for SLMs and small vision models employing the 1.58-bit does not follow the findings in prior work \cite{ma2024era} to employ significantly larger learning-rates, even when trained from scratch. Weight decay distorts the training when employed in training with a high learning rate, but to the contrary helps when applied with smaller learning rates.
% mean /median
Our results on employing AbsMean vs AbsMedian quantization of the 16-bit shadow weights do not yield distinctive and conclusive results, leaving it as a hyperparameter for now and opening avenues for future work on the most advantageous quantization schemes from 16 to 1.58 bits in the context of quantization-aware training.

\bibliographystyle{splncs04}
\bibliography{sample.bib}





% LNCS Homepage, \url{http://www.springer.com/lncs}, last accessed 2023/10/25
% \end{thebibliography}
\end{document}